\documentclass[10pt,twocolumn,letterpaper]{article}

\usepackage[pagenumbers]{cvpr} %

\usepackage{marvosym}

\usepackage{amsthm}      
\usepackage{multirow}
\usepackage{tocloft}
\usepackage{makecell}
\usepackage{setspace}
\definecolor{mygray}{gray}{.92}
\usepackage[table]{xcolor} 
\usepackage{etoolbox}
\usepackage{listings}
\newlength\savewidth
    
\newtheorem*{theorem*}{Theorem} 

\definecolor{cvprblue}{rgb}{0.21,0.49,0.74}
\usepackage[pagebackref,breaklinks,colorlinks,allcolors=cvprblue]{hyperref}
\usepackage{float}

\def\paperID{****} %
\def\confName{CVPR}
\def\confYear{2026}

\title{ApET: Approximation-Error Guided Token Compression for Efficient VLMs}

\author{
  Qiankun Ma$^{1, 2, 3,*}$ \quad
  Ziyao Zhang$^{1, 2, 3,*}$ \quad
  Haofei Wang$^{2}$ \quad 
  Jie Chen$^{2,4,5}$ \\
  Zhen Song$^{2}$ \quad
  Hairong Zheng$^{1,2,3, \dagger}$
  \vspace{2mm} \\
  $^1$Shenzhen Institutes of Advanced Technology, Chinese Academy of Sciences \quad $^2$Peng Cheng Laboratory \\ 
  \quad $^3$University of Chinese Academy of Sciences \quad $^4$Harbin Institute of Technology \quad $^5$Peking University 
}

\renewcommand{\footnoterule}{%
  \kern -3pt
  \hrule width 0.9\linewidth height 0.4pt
  \kern 2.6pt
}

\usepackage{standalone}
\usepackage{pgfplots}
\usepgfplotslibrary{polar}
\pgfplotsset{compat=1.18}

\begin{document}
\maketitle

\begingroup
\renewcommand\thefootnote{}
\footnotetext{
  $^*$Equal contributions.
  $^\dagger$Corresponding author.
}
\endgroup

\begin{abstract}
Recent Vision-Language Models (VLMs) have demonstrated remarkable multimodal understanding capabilities, yet the redundant visual tokens incur prohibitive computational overhead and degrade inference efficiency. Prior studies typically relies on [CLS] attention or text-vision cross-attention to identify and discard redundant visual tokens. Despite promising results, such solutions are prone to introduce positional bias and, more critically, are incompatible with efficient attention kernels such as FlashAttention, limiting their practical deployment for VLM acceleration. In this paper, we step away from attention dependencies and revisit visual token compression from an information-theoretic perspective, aiming to maximally preserve visual information without any attention involvement. We present ApET, an \textbf{Ap}proximation-\textbf{E}rror guided \textbf{T}oken compression framework. ApET first reconstructs the original visual tokens with a small set of basis tokens via linear approximation, then leverages the approximation error to identify and drop the least informative tokens. Extensive experiments across multiple VLMs and benchmarks demonstrate that ApET retains 95.2\% of the original performance on image-understanding tasks and even attains 100.4\% on video-understanding tasks, while compressing the token budgets by 88.9\% and 87.5\%, respectively. Thanks to its attention-free design, ApET seamlessly integrates with FlashAttention, enabling further inference acceleration and making VLM deployment more practical.
Code is available at \href{https://github.com/MaQianKun0/ApET}{{\texttt{https://github.com/MaQianKun0/ApET}}}.
\end{abstract}
    
\section{Introduction}
\label{sec:intro}

Recent advances in Vision-Language Models (VLMs)~\cite{bai2023qwen,chen2024sharegpt4v,li2023blip,li2024mini,liu2024improved,chen2024longvila,jia2024leopard,li2024llava,liu2024llavanext,ma2024vista,maaz2023video, lu2025mambatad, lu2025pretrain} have demonstrated their exceptional capabilities in understanding and reasoning about visual content across a wide range of vision-language tasks. However, VLMs face significant challenges in computational efficiency and scalability, primarily due to the large number of visual tokens required to represent high-resolution images and long video sequences. These burdensome demands markedly constrain the practical deployment of VLMs in real-world scenarios~\cite{kim2024openvla,qu2025mobile,yang2025lidar,yang2024unified,yao2024minicpm}.

\begin{figure}[t]
    \centering
    \includegraphics[width=1\linewidth]{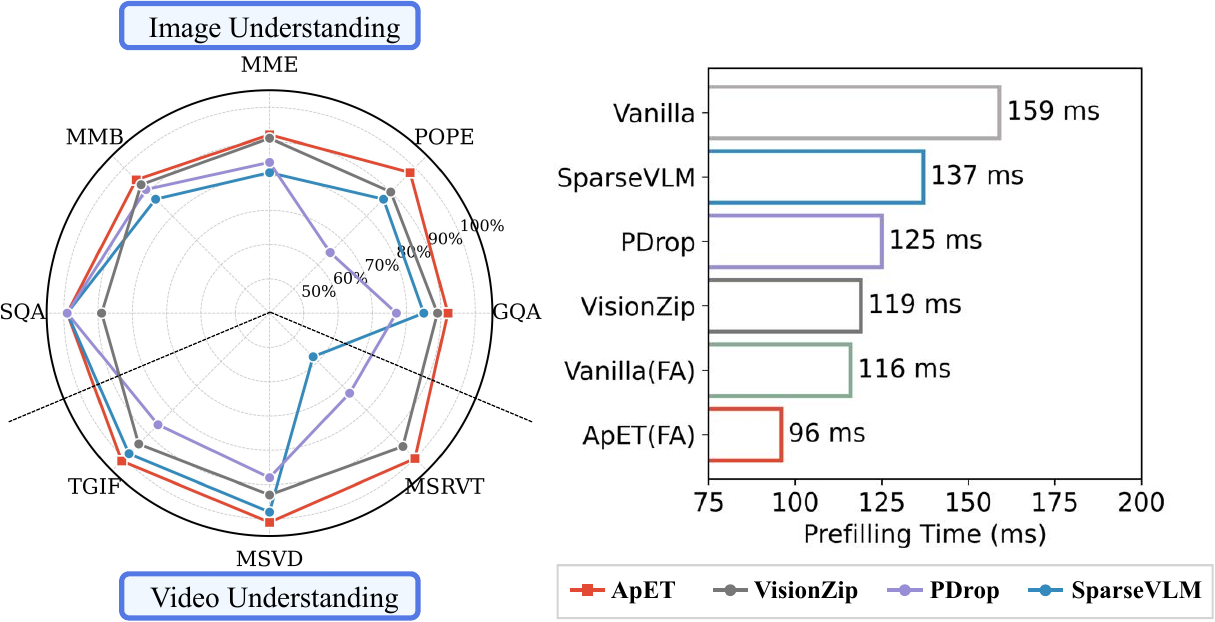}
    \caption{\textbf{Performance–Efficiency Comparison.} (Left) Performance across various image- and video-understanding benchmarks, ApET significantly outperforms existing token reduction approaches. (Right) ApET can be seamlessly combined with FlashAttention (FA) to further reduce prefilling time on Qwen-2.5-VL, whereas prior token reduction approaches are incompatible with FlashAttention.
    }
    \label{fig:teaser-performance}
    \vspace{-0.4cm}
\end{figure}

Recognizing that many visual tokens are uninformative for VLM response, recent studies~\cite{chen2024image,xing2024pyramiddrop,yang2025visionzip,zhang2024sparsevlm,zhang2025vscan,xu2025rethinking} have introduced training-free token compression techniques that prune or merge less-informative tokens to improve efficiency without sacrificing performance. These methods typically use attention weights to identify and retain salient tokens. However, their inherent dependence on the attention mechanism still limits their practical implementation. Specifically, since these methods require access to attention weights to determine token importance, they are incompatible with efficient attention implementations such as FlashAttention (FA), which do not provide attention weights. Consequently, such compression techniques cannot be seamlessly integrated with these optimized attention operators.

\begin{figure}[t]
    \centering
    \includegraphics[width=1\linewidth]{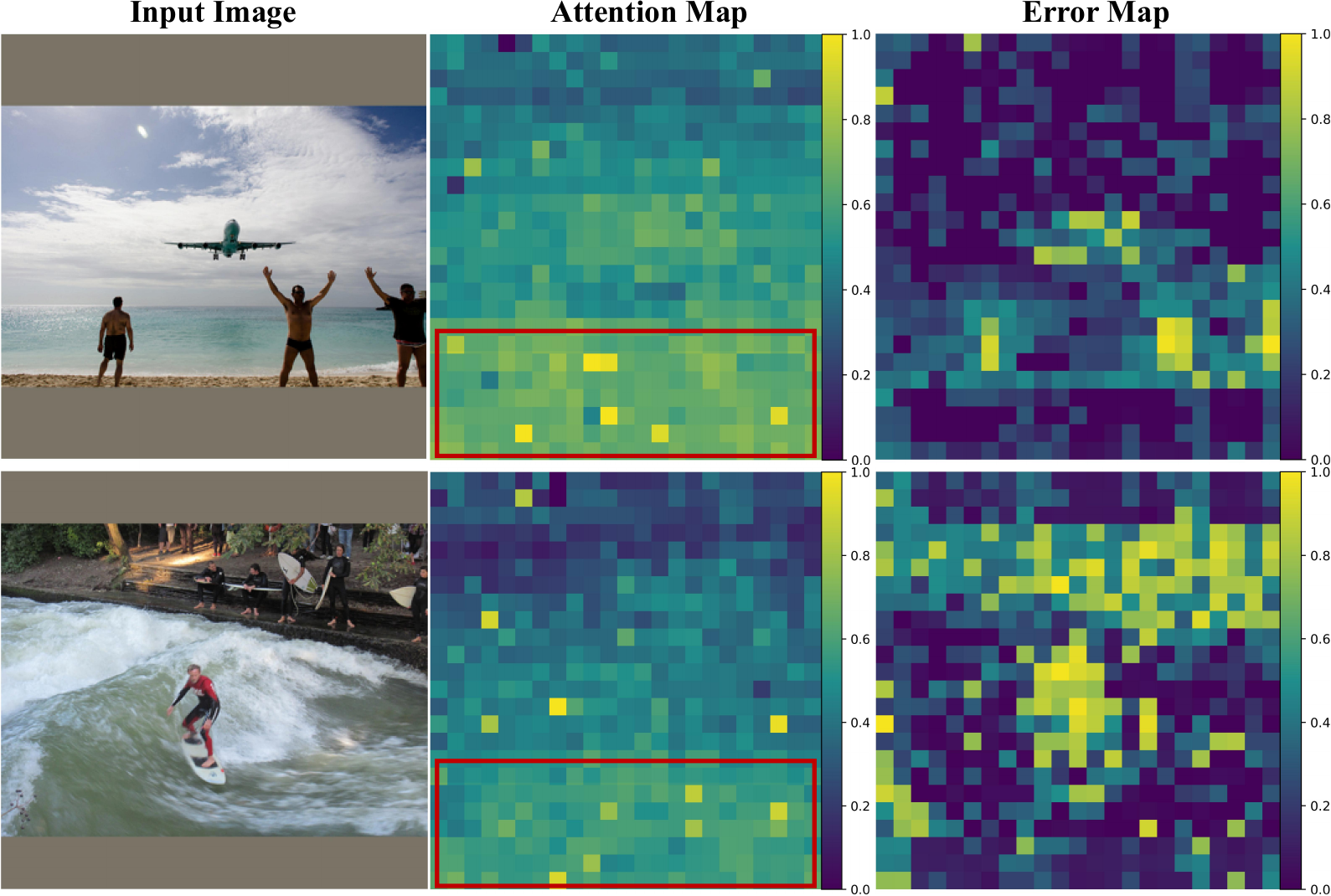}
    \caption{\textbf{Comparison of the attention map and approximate error map.} Attention-guided methods exhibit a positional bias that is agnostic to actual information content, assigning disproportionately high importance to later positions regardless of their semantic relevance (red boxes). In contrast, the error map provides an intuitive and content-aware reflection of token importance.
    }
    \label{fig:visual}
    \vspace{-0.4cm}
\end{figure}

To empirically demonstrate this, we conducted an efficiency comparison experiment (\figureautorefname~\ref{fig:teaser-performance} right). The results reveal that existing attention-guided compression methods are less efficient than directly employing FA. Moreover, since the use of FA incurs virtually no degradation in model performance, attention-guided compression approaches offer no practical advantage in real-world applications. Additionally, the visualization in \figureautorefname~\ref{fig:visual} reveals a positional bias inherent in the attention mechanisms of LLMs: visual tokens positioned closer to textual tokens (i.e., those appearing later in the sequence) consistently receive disproportionately high attention. This bias increases the risk that potentially important tokens will be erroneously discarded during compression, thereby degrading model performance.

To address the aforementioned challenges, we propose the \textbf{Ap}proximation-\textbf{E}rror guided \textbf{T}oken reduction method (ApET). ApET first linearly reconstructs each token using a carefully selected subset of tokens and then estimates its information content by computing the reconstruction error. This enables us to evaluate token importance without relying on external signals. Our key insight is that a token’s intrinsic information can be effectively captured by its linear approximation error: a small reconstruction error indicates that the token can be well represented by the selected subset, suggesting low information content, whereas a large error implies higher information content (see Section~\ref{sec:method} for details). Unlike attention-guided methods, ApET focuses on the intrinsic characteristics of tokens, thereby inherently avoiding position bias, as illustrated in \figureautorefname~\ref{fig:visual}. Furthermore, ApET not only achieves state-of-the-art performance on both image- and video-understanding tasks but also remains compatible with efficient attention kernels such as FA, enabling further computational efficiency, as demonstrated in \figureautorefname~\ref{fig:teaser-performance}.

\noindent We summarize our contributions as follows:
\begin{itemize}
\itemsep=-1pt
    \item We conduct the first comprehensive analysis of visual token evaluation in VLMs from an information-theoretic perspective. By estimating the information carried by each token via its approximation error, we can assess token importance without relying on external signals, yielding a novel perspective on visual token compression in VLMs.

    \item Based on our exploration, we introduce ApET, an approximation-error guided token compression method that reconstructs tokens via linear approximation and quantifies their importance by the approximation-error, thereby eliminating positional bias while retaining compatibility with efficient operators.

    \item Extensive experiments show that ApET achieves outstanding performance on both image- and video-understanding benchmarks, while its fusion with FlashAttention yields an additional gain in model efficiency.
\end{itemize}

\section{Related Work}
\label{sec:related_work}
\textbf{Efficient Vision-Language Models. }
Leveraging powerful auto-regressive LLMs~\cite{achiam2023gpt,chung2024scaling,touvron2023llama,liu2024deepseek}, contemporary VLMs predominantly adopt an encoder-projector-decoder paradigm that fuses visual tokens with textual sequences~\cite{chen2023shikra,chen2024internvl,team2024gemini,tong2024cambrian}. However, as image resolution or the number of frames increases, the volume of visual tokens increases proportionally, triggering a quadratic increase in computational cost due to the self-attention mechanism~\cite{brauwers2021general,child2019generating,vaswani2017attention,zaheer2020big} and hampering deployment in resource-constrained scenarios~\cite{chen2024image,bolya2022token,kondratyuk2021movinets,mehta2021mobilevit,zhang2025self,zhang2025vlm2}. To counteract this limitation, recent models embed specialized bottlenecks—e.g., InstructBLIP’s Q-Former~\cite{dai2023instructblip} and OpenFlamingo’s perceiver resampler~\cite{jaegle2021perceiver,alayrac2022flamingo} —that compress dense visual representations into compact feature sets before reaching the LLM decoder.  Complementing these architectural refinements, FlashAttention~\cite{dao2022flashattention,dao2023flashattention} reorders memory access patterns to exploit GPU hierarchies, yielding substantial speed-ups without sacrificing downstream accuracy.

\noindent \textbf{Visual Token Compression for VLMs. }
To mitigate the quadratic complexity of self-attention introduced by dense vision tokens, existing approaches can be broadly categorized into two types: 1) \textit{Text-guided compression methods}~\cite{liu2024multi,tan2025tokencarve,xing2025conical,ye2025atp,zhang2024sparsevlm}, which leverage textual queries to eliminate visual tokens that are irrelevant to the text. For example, SparseVLM~\cite{zhang2024sparsevlm} introduces an iterative sparsification strategy that selects visually relevant text tokens to assess the importance of vision tokens. Similarly, PyramidDrop~\cite{xing2025conical} performs progressive pruning across multiple decoding layers to balance computational efficiency with contextual integrity. 2) \textit{Text-agnostic compression methods}~\cite{arif2025hired,shang2024llava,wang2025folder,wen2025stop,yang2025visionzip,zhang2024cls}, which rely solely on self-attention among visual tokens and the [CLS] token to identify and prune redundant visual information. For instance, VisionZip~\cite{yang2025visionzip} selects dominant tokens based on attention scores associated with the [CLS] token, while VisionDrop~\cite{xu2025rethinking} filters significant tokens using self-attention weights between visual tokens. Despite their effectiveness, these methods typically depend on attention weights, which can conflict with the implementation of efficient attention mechanisms such as FlashAttention. In this work, we propose an information-theoretic approach that removes the least informative tokens without relying on attention weights. This ensures full compatibility with FlashAttention while achieving compression performance on par with existing methods.
\section{Method}
\subsection{Preliminary}
Most existing VLMs are composed of three core components: a visual encoder, a feature projector, and an LLM decoder. Taking an image as input, the visual encoder first processes the image patches, and the projector converts them into a set of $n$ visual tokens, denoted as $V=\{v_1,\ldots,v_n\}$. These visual tokens are then prepended to the tokenized textual query $T$ and fed into the LLM decoder for auto-regressive next-token generation. This process can be formally expressed as: $y_t\sim p_{\theta}(y_t |V,T, \mathbf{y}_{<t})$, where $\theta$ denotes the parameters of the VLM, $y_t$ is the next token sampled from the output probability distribution $p_{\theta}(\cdot)$, and $\mathbf{y}_{<t}$ represents the sequence of tokens generated before timestep $t$.

\begin{figure*}[t]
    \centering
    \includegraphics[width=0.95\linewidth]{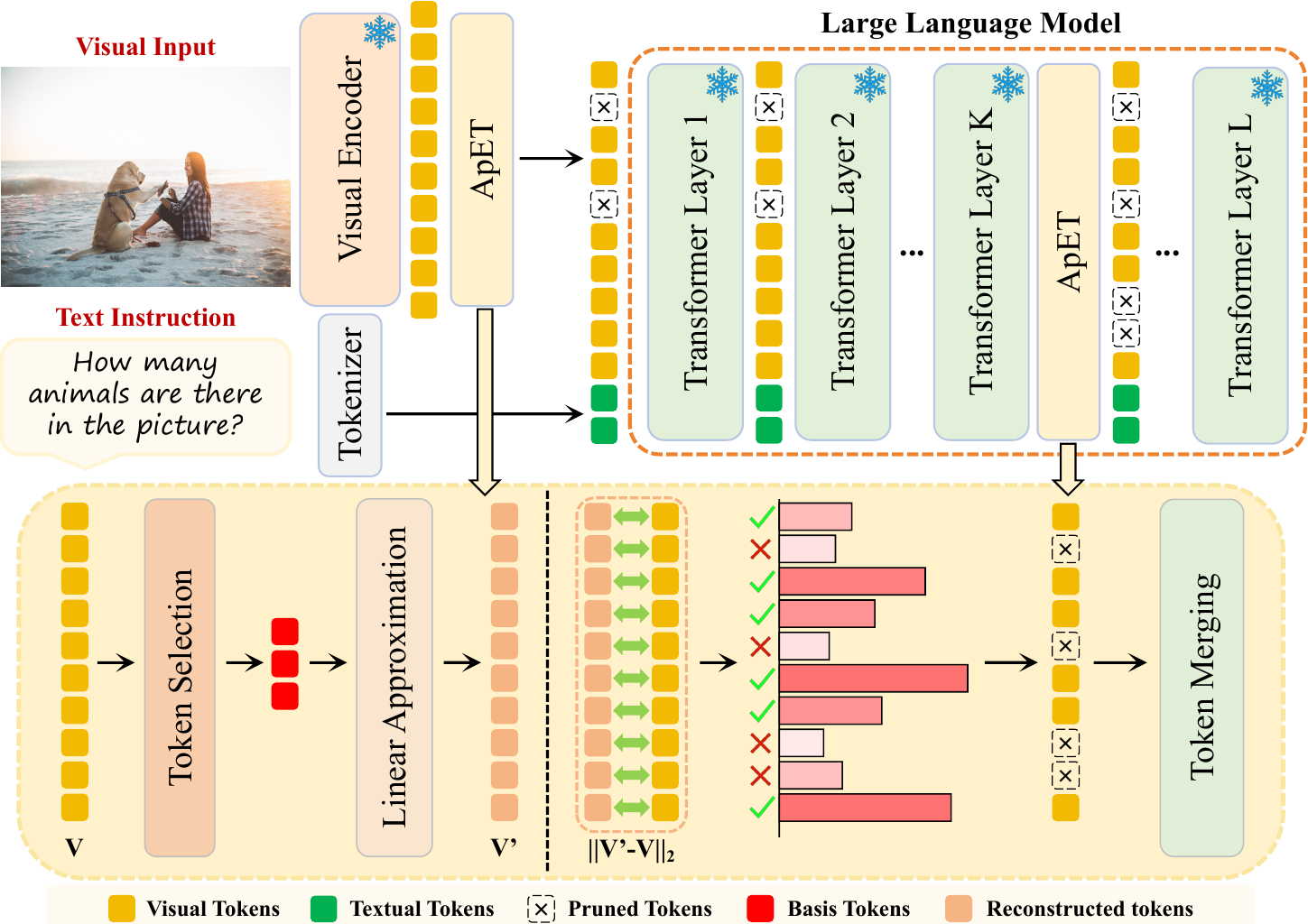}
     % \vspace{-0.2cm}
\caption{\textbf{The overview of ApET.} ApET selectively filters tokens exhibiting minimal information content based on approximation error, and subsequently employs token merging strategies to accomplish token compression.}
    \label{fig:framework}
\end{figure*}

\subsection{Information-Theoretic Analysis}
\label{sec:method}
Unlike most previous approaches that rely on attention weights to assess token importance, we adopt an information-theoretic perspective, aiming to maximize the information content retained in compressed visual tokens. Specifically, given a set of visual tokens $V$, we sample a compressed subset $S$ with the objective of maximizing the mutual information $I(V;S)$, as follows:
\begin{equation}
\label{eq:max_mi}
S^{*}=
\arg\max_{\substack{S}}
I(V;S)
\end{equation}
The objective is to identify an optimal subset $S \subseteq V$ that retains the maximum information about the original set $V$. According to information theory~\cite{shannon1948mathematical}, the mutual information $I(V;S)$ can be defined as:
\begin{equation}
\label{eq:mi}
I(V;S)=H(V)-H(V | S)
\end{equation}
Here, $H(V)$ denotes the entropy of $V$, while $H(V | S)$ quantifies the remaining uncertainty after observing $S$. Since $H(V)$ is constant for a given input, maximizing $I(V;S)$ reduces to minimizing $H(V | S)$. In practice, however, this minimization is intractable: the joint distribution $p(V,S)$ required to compute $H(V | S)$ is unavailable.
\vspace{-1mm}
\begin{theorem*}
\textbf{(Lower bound on the minimal reconstruction MSE $\xi$).}
Let $H(x|z)$ denote the conditional entropy of the input $x$ given the intermediate feature $z$. The minimal reconstruction MSE $\xi$ is bounded by: $\xi \ge \frac{1}{(2\pi e)}exp(\frac{2H({x}|{z})}{d})$.
\label{theorem1}
\end{theorem*}
\vspace{-5mm}
\noindent Leveraging this theorem, we can bound the conditional entropy in terms of the expected reconstruction mean-square error (MSE)~\cite{xia2025theoretical}. Specializing the general bound to our setting yields:
\begin{equation}
\label{eq:theorem}
\frac{1}{(2\pi e)}exp(\frac{2H({V}|{S})}{d}) \le \xi
\end{equation}
According to Eq.~\ref{eq:theorem}, minimising the reconstruction MSE $\xi$ is connected to minimizing the conditional entropy $H({V}|{S})$. Thus, the significance of visual tokens can be evaluated by quantifying the reconstruction error.

\subsection{Approximation-Error Token Compression}
Based on our analysis, reconstruction MSE serves as an effective indicator of the importance of visual tokens—larger errors correspond to greater token significance. To ensure simplicity and computational efficiency, we eschew the use of an additional reconstruction model. Instead, we adopt a linear approximation strategy, leveraging the resulting approximation error to evaluate token importance. 
Specifically, we first select a compact subset of tokens as base tokens and linearly reconstruct the remaining tokens from them. We then compute the per-token feature discrepancy between the original and reconstructed representations. Grounded in the preceding theoretical analysis, we adopt the magnitude of this reconstruction error as a principled surrogate for token importance.
Building upon this, we propose the \textbf{Ap}proximation-\textbf{E}rror guided \textbf{T}oken compression method, as illustrated in \figureautorefname~\ref{fig:framework}.

ApET can be seamlessly integrated at any layer of either the vision encoder or the LLM. In this work, we instantiate it immediately after the vision encoder and at an intermediate LLM layer (e.g., layer 16 in LLaVA). Concretely, ApET proceeds in three stages: token selection, ranking by approximation-error, and token merging.

\noindent \textbf{Token Selection. } Given a visual token set $V^l$ at layer $l$, we construct $M$ basis token subset $B=\{b_1,\ldots,b_M\}$ via sampling, where the hyper-parameter $M$ controls the sampling size. Three sampling strategies are investigated in thie paper: 1) Farthest Distance Sampling (FPS), 2) Density Peak Clustering (DPC), and 3) Random Sampling. Ablation results are reported in Table~\ref{tab:ab_sampling}.

\noindent \textbf{Approximation-Error Computation. } For each token $v \in V$, we seek a linear combination of the tokens in $B$ that approximates $v$:
\begin{equation}
\label{eq:linear_ap}
v' \approx \sum_{i=1}^{M} \alpha_i b_i
\end{equation}
Here, the coefficients $\alpha_i$ are assembled into a matrix $A \in \mathbb{R}^{M \times N}$, so that the entire set $V$ is approximated by the single relation:$V \approx  BA$. This coefficient matrix $A$ is obtained by solving the corresponding linear system. After that, the approximate error used to measure the information of a token is defined as:
\begin{equation}
\label{eq:error}
\xi =\Delta v= ||v-v'||_2
\end{equation}
The approximation error can be employed to evaluate token significance while remaining compatible with FA.

\noindent \textbf{Token Merging. } Based on approximation error, we rank the importance of visual tokens and selectively reduce those with lower significance. It is worth noting that, in order to avoid omitting any potentially important tokens, we explicitly retain the base tokens as part of the significant token set. To mitigate potential information loss, we propose a similarity-based token merging strategy. Specifically, for each token marked for removal, we identify and merge it with the most similar token among those retained. After all unselected tokens have been assigned to their closest counterparts, we apply average merging~\cite{bolya2022token} within each group to generate merged tokens.
\section{Experiments}
\renewcommand{\multirowsetup}{\centering}
	\definecolor{mygray}{gray}{.92}
	\definecolor{ForestGreen}{RGB}{34,139,34}
	\newcommand{\fg}[1]{\mathbf{\mathcolor{ForestGreen}{#1}}}
	\definecolor{Forestred}{RGB}{220,50,50}
	\definecolor{lightgreen}{rgb}{0.886, 0.941, 0.851}
	\newcommand{\fr}[1]{\mathbf{\mathcolor{Forestred}{#1}}}
	\begin{table*}[t]
		\centering
        \caption{\textbf{Performance of ApET on LLaVA-1.5 across 9 image understanding benchmarks.} The original number of visual tokens is 576. The last column reports the average accuracy relative to the theoretical upper bound. The best result in each setting is highlighted.}
		\label{tab:main_llava1.5}
		\setstretch{1.3}
		\setlength{\tabcolsep}{4pt}
        \footnotesize
		\resizebox{0.95\linewidth}{!}{
			\begin{tabular}{l | c c c c c c c c c c c c| >{\centering\arraybackslash}p{1.2cm}}
				\toprule[1.2pt]
				\footnotesize\textbf{Method} & \footnotesize\textbf{ GQA } & \footnotesize\textbf{MMB} & \footnotesize\textbf{MMB}$^{\text{CN}}$ & \footnotesize\textbf{MME} & \footnotesize\textbf{ POPE } & \footnotesize\textbf{ SQA } & \footnotesize\textbf{VQA}$^{\text{V2}}$ & \footnotesize\textbf{VQA}$^{\text{Text}}$ & \footnotesize\textbf{VizWiz} &  \makecell[c]{\textbf{Avg}.}\\
				\hline
				
				\rowcolor{mygray}
				\multicolumn{11}{c}{\textit{Upper Bound, 576 Tokens} \ $\textbf{(100\%)}$}\\
				\textcolor{gray}{Vanilla} & \textcolor{gray}{61.9} & \textcolor{gray}{64.7} & \textcolor{gray}{58.1} & \textcolor{gray}{1862} & \textcolor{gray}{85.9} & \textcolor{gray}{69.5} & \textcolor{gray}{78.5} & \textcolor{gray}{58.2}  & \textcolor{gray}{50.0} & \multirow{1}*{\textcolor{gray}{100\%}} \\
				\hline
				
				\rowcolor{mygray}
				\multicolumn{11}{c}{\textit{Retain 192 Tokens} \ $\fg{(\downarrow 66.7\%)}$}\\
				ToMe \texttt{\scriptsize{(ICLR23)}} & 54.3 & 60.5 & - & 1563 & 72.4 & 65.2 & 68.0 & 52.1 & - &88.5\% \\
				FastV \texttt{\scriptsize{(ECCV24)}} & 52.7 & 61.2 & 57.0 & 1612 & 64.8 & 67.3 & 67.1 & 52.5 & 50.8 &90.4\% \\
				SparseVLM \texttt{\scriptsize{(ICML25)}} & 57.6 & 62.5 & 53.7 & 1721 & 83.6 & \textbf{69.1} & 75.6 & 56.1 & 50.5 & 96.1\% \\
				PDrop \texttt{\scriptsize{(CVPR25)}} & 57.3 & 63.3 & 56.8 & 1797 & 82.3 & 69.0 & 75.1 & 56.5 & \textbf{51.1} & 97.2\% \\ 
				VisionZip \texttt{\scriptsize{(CVPR25)}} & 59.3 & 63.0 & - & 1783 & 85.3 & 68.9 & \textbf{77.4} & \textbf{57.3} & - & 97.8\%    \\
				
				\rowcolor{blue!12}
				\textbf{ApET (Ours)} & \textbf{60.2} & \textbf{63.4} & \textbf{57.9} & \textbf{1808} &\textbf{86.3} & 68.5 & 76.2 & 54.4 & 50.2 & \textbf{98.0\%} \\
				\hline
				
				\rowcolor{mygray}
				\multicolumn{11}{c}{\textit{Retain 128 Tokens} \ $\fg{(\downarrow 77.8\%)}$}\\
				ToMe \texttt{\scriptsize{(ICLR23)}} & 52.4 & 53.3 & - & 1343 & 62.8 & 59.6 & 63.0 & 49.1 & - & \multirow{1}*{80.4\%} \\
				FastV \texttt{\scriptsize{(ECCV24)}} & 49.6 & 56.1 & 56.4 & 1490 & 59.6 & 60.2 & 61.8 & 50.6 & 51.3 & \multirow{1}*{85.4\%}\\
				SparseVLM \texttt{\scriptsize{(ICML25)}} & 56.0 & 60.0 & 51.1 & 1696 & 80.5 & 67.1 & 73.8 & 54.9 & \textbf{51.4} & 93.7\% \\
				PDrop \texttt{\scriptsize{(CVPR25)}} & 57.1 & 61.6 & \textbf{56.6} & 1761 & 82.3 & 68.4 & 72.9 & 56.6 & 51.0 & 96.2\% \\
				VisionZip \texttt{\scriptsize{(CVPR25)}} & 57.6 & 62.0 & - & 1763 & 83.2 & \textbf{68.9} & \textbf{75.6} & \textbf{56.8} & - & 96.2\%    \\
				
				\rowcolor{blue!12}
				\textbf{ApET (Ours}) & \textbf{58.9} & \textbf{62.3} & 56.4 & \textbf{1801} & \textbf{86.1} & 68.7 & 75.1 & 53.9 & 50.7 & \textbf{97.1\%} \\
				\hline
				
				\rowcolor{mygray}
				\multicolumn{11}{c}{\textit{Retain 64 Tokens} \ $\fg{(\downarrow 88.9\%)}$}\\
				ToMe \texttt{\scriptsize{(ICLR23)}} & 48.6 & 43.7 & - & 1138 & 52.5 & 50.0 & 57.1 & 45.3 & - & 70.1\%\\
				FastV \texttt{\scriptsize{(ECCV24)}} & 46.1 & 48.0 & 52.7 & 1256 & 48.0 & 51.1 & 55.0 & 47.8 & 50.8 & 76.7\% \\
				SparseVLM \texttt{\scriptsize{(ICML25)}} & 52.7 & 56.2 & 46.1 & 1505 & 75.1 & 62.2 & 68.2 & 51.8 & 50.1 & 87.2\% \\
				PDrop \texttt{\scriptsize{(CVPR25)}} & 47.5 & 58.8 & 50.5 & 1561 & 55.9 & \textbf{69.2} & 69.2 & 50.6 & 50.7 & 86.6\% \\
				VisionZip \texttt{\scriptsize{(CVPR25)}} & 55.1 & 60.1 & - & 1690 & 77.0 & 69.0 & 72.4 & \textbf{55.5} & - & 92.7\%    \\
				
				\rowcolor{blue!12}
				\textbf{ApET (Ours}) & \textbf{56.9} & \textbf{61.2} & \textbf{54.4} & \textbf{1714} & \textbf{84.4} & 68.9 & \textbf{72.5} & 53.0 & \textbf{51.9} & \textbf{95.2\%} \\
				\bottomrule[1.2pt]
			\end{tabular}
		}
	\end{table*}
    
\subsection{Experimental Settings}
\textbf{Models and Baselines.} To validate the effectiveness of our proposed method, we apply ApET to four representative VLMs. For image understanding tasks, we integrate our approach into the widely adopted LLaVA family, specifically LLaVA-1.5-7B~\cite{liu2024improved} and LLaVA-Next-7B~\cite{liu2024llavanext}. To further assess the generalizability of our method, we extend the evaluation to the more advanced Qwen2.5-VL-7B model~\cite{bai2025qwen2}. In the context of video understanding, we adopt Video-LLaVA-7B~\cite{lin2023video} as the baseline model and evaluate our approach on three commonly used video benchmarks. We compare the performance of ApET against several state-of-the-art visual token reduction methods, including ToMe~\cite{bolya2022token}, FastV~\cite{chen2024image}, SparseVLM~\cite{zhang2024sparsevlm}, PDrop~\cite{xing2025conical}, and VisionZip~\cite{yang2025visionzip}.

\noindent \textbf{Implementation Details. } We adopt the standard inference configurations provided in the official implementations of each VLM. Our method can be seamlessly integrated into these models in a training-free manner. Specifically, for each VLM, we perform visual token compression at the output of the visual encoder and within a middle layer of the LLM: layer 16 for the LLaVA-series models and layer 14 for Qwen2.5-VL.

\begin{table}[t]
		\centering
        \caption{\textbf{Performance comparison of ApET on LLaVA-NeXT-7B across multiple image understanding benchmarks.}}
		\label{tab:main_llava_next}
		\setstretch{1.3}
		\setlength{\tabcolsep}{0.5pt}
        \small
		\resizebox{1.0\linewidth}{!}{
			\begin{tabular}{l | c c c c c c c c c c| >{\centering\arraybackslash}p{1.2cm}}
				\toprule[1.2pt]
				\small\textbf{Method} & \small\textbf{ GQA } & \small\textbf{MMB} & \small\textbf{MMB}$^{\text{CN}}$ & \small\textbf{MME} & \small\textbf{ POPE }  & \small\textbf{VQA}$^{\text{V2}}$ & \small\textbf{VQA}$^{\text{Text}}$ &  \makecell[c]{\textbf{Avg}.}\\
				\hline
				
				\rowcolor{mygray}
				\multicolumn{9}{c}{\textit{Upper Bound, 2880 Tokens} \ $\textbf{(100\%)}$}\\
				\textcolor{gray}{Vanilla} & \textcolor{gray}{64.2} & \textcolor{gray}{67.4} & \textcolor{gray}{60.6} & \textcolor{gray}{1851} & \textcolor{gray}{86.5} & \textcolor{gray}{81.8} & \textcolor{gray}{61.3}  & \multirow{1}*{\textcolor{gray}{100\%}} \\
				\hline

                \rowcolor{mygray}
				\multicolumn{9}{c}{\textit{Retain 640 Tokens} \ $\fg{(\downarrow 77.8\%)}$}\\
				SparseVLM  & 60.3 & 65.8 & 58.5 & 1773 & 84.2 & 77.1 & 57.8 & 95.7\% \\
				PDrop  & 60.6 & 65.5 & 58.5 & 1781 & 83.7  & 78.3 & 57.4 & 95.8\% \\ 
				VisionZip  & 61.3 & \textbf{66.2} & - & 1787 & 85.9 & 79.1 & \textbf{60.2} & 97.4\%    \\
				
				\rowcolor{blue!12}
				\textbf{ApET (Ours)} & \textbf{63.0} & 65.3 & \textbf{59.3} & \textbf{1815} &\textbf{87.2} & \textbf{79.2} & 57.9 & \textbf{97.6\%} \\
				\hline
				
				\rowcolor{mygray}
				\multicolumn{9}{c}{\textit{Retain 320 Tokens} \ $\fg{(\downarrow 88.9\%)}$}\\
				SparseVLM  & 59.3 & \textbf{64.2} & 55.9 & 1690 & 83.3 & 75.7 & 58.8 & 93.7\% \\
				PDrop  & 56.4 & 63.4 & 56.2 & 1663 & 77.6  & 73.5 & 54.4 & 90.4\% \\ 
				VisionZip  & 59.3 & 63.1 & - & 1702 & 82.1 & \textbf{76.2} & \textbf{58.9} & 93.7\%    \\
				
				\rowcolor{blue!12}
				\textbf{ApET (Ours)} & \textbf{61.0} & 63.5 & \textbf{56.6} & \textbf{1783} &\textbf{85.6} & 75.8 & 54.4 & \textbf{94.2\%} \\
				\hline
				
				\rowcolor{mygray}
				\multicolumn{9}{c}{\textit{Retain 160 Tokens} \ $\fg{(\downarrow 94.4\%)}$}\\
				SparseVLM  & 51.2 & 52.1 & 48.6 & 1542 & 72.7 & 66.3 & 46.4 & 80.2\% \\
				PDrop  & 54.9 & \textbf{61.8} & \textbf{54.9} & 1513 & 72.3 & 70.2 & 52.7 & 86.4\% \\
				VisionZip  & 55.5 & 60.1 & - & 1628 & 74.8 & 71.4 & \textbf{56.2} & 88.2\%    \\
				
				\rowcolor{blue!12}
				\textbf{ApET (Ours}) & \textbf{58.4} & 60.8 & 52.3 & \textbf{1680} & \textbf{82.6} & \textbf{72.7} & 53.8 & \textbf{90.1\%} \\
				\bottomrule[1.2pt]
			\end{tabular}
		}
	\end{table}

\subsection{Main Results}
\textbf{Image Understanding Tasks. } As shown in Table~\ref{tab:main_llava1.5}, we integrate ApET to LLaVA-1.5-7B and compare its performance with existing baselines across 9 image understanding tasks. Following the setup
in~\cite{yang2025visionzip,zhang2024sparsevlm}, we compress the visual token sequence to 192, 128 and 64 tokens to evaluate the advantages of our method. Our method consistently outperforms all competing methods across different token reduction rates, with its superiority becoming increasingly pronounced at higher compression ratios. Specifically, even under an aggressive reduction rate of 88.9\%, it's average performance outperforms the second-best VisionZip by a substantial margin of 2.5\%. 

To further validate the effectiveness of our approach, we extend our evaluation to LLaVA-NeXT-7B. Unlike LLaVA-1.5, LLaVA-NeXT decomposes each image into four parts together with a down-scaled full-view, yielding five distinct images and consequently a markedly larger visual token budget, making visual token compression more imperative. Following the experimental protocol in~\cite{yang2025visionzip}, we conducted comparative experiments under three vision token budgets (640, 320, and 160). As shown in Table~\ref{tab:main_llava_next}, our method consistently exhibits superior performance across all three configurations. Specifically, using only 160 tokens, our method achieves 97.6\% average accuracy, attaining the best performance on 4 out of 7 benchmarks.

Furthermore, we evaluate the competing algorithms on the more capable model, that is, Qwen-2.5-VL-7B. Unlike previous work, it adopts dynamic resolution, so each image is encoded into a variable length sequence of visual tokens. The length is bounded by the min pixels and max pixels hyper-parameters, which we fix to 256 and 2 048, respectively, for all methods to ensure a fair comparison.
Table~\ref{tab:main_qwen} reports the results on five image-understanding benchmarks. ApET consistently surpasses all baseline at every retention rate, confirming its efficacy at high resolution and its seamless compatibility with variable-resolution inputs.

Notably, most baselines estimate token importance based on attention weights. Although this heuristic is simple and often effective, its weight-access pattern is fundamentally incompatible with FlashAttention. Our method, by contrast, not only matches or exceeds their accuracy, but also plugs seamlessly into FlashAttention, delivering both efficiency and scalability.

\begin{table}[t]
		\centering
        \caption{\textbf{Performance comparisons on Qwen-2.5-VL-7B across five image understanding benchmarks.} \dag Evaluation is based on our re-implementation.}
		\label{tab:main_qwen}
		\setstretch{1.3}
		\setlength{\tabcolsep}{5pt}
        \small
		\resizebox{1.0\linewidth}{!}{
			\begin{tabular}{l| c c c c c c >{\centering\arraybackslash}p{1.2cm}}
				\toprule[1.2pt]
				\small\textbf{Method} & \small\textbf{ GQA } &\small\textbf{POPE } & \small\textbf{SQA}  &  \small\textbf{MME} & \small\textbf{MMB} &  \makecell[c]{\textbf{Avg}.}\\
				\hline
				
				\rowcolor{mygray}
				\multicolumn{7}{c}{\textit{Upper Bound, 256$\sim$2048 Tokens} \ $\textbf{(100\%)}$}\\
				\textcolor{gray}{Vanilla} & \textcolor{gray}{60.5} &  \textcolor{gray}{86.2} & \textcolor{gray}{76.7} & \textcolor{gray}{2327} & \textcolor{gray}{83.3}  & \multirow{1}*{\textcolor{gray}{100\%}} \\
				\hline

                \rowcolor{mygray}
				\multicolumn{7}{c}{\textit{Retain 20\% Tokens in Average} \ $\fg{(\downarrow 80\%)}$}\\
				SparseVLM\dag  & 54.7 & 73.6 & 71.6 & 2063 & 76.0 & 89.8\% \\
				PDrop\dag  & 55.1 & 78.4 & 70.9  & 2117 & 77.3 & 91.6\% \\ 
				VisionZip\dag   & 56.8 & 82.4 & \textbf{76.3} & 2134 & \textbf{79.3} & 95.2\%   \\
				
				\rowcolor{blue!12}
				\textbf{ApET (Ours)} & \textbf{58.8} & \textbf{85.1} & 74.9 & \textbf{2181} &78.6 & \textbf{96.3\%} \\
				\hline
				
				\rowcolor{mygray}
				\multicolumn{7}{c}{\textit{Retain 10\% Tokens in Average} \ $\fg{(\downarrow 90\%)}$}\\
				SparseVLM\dag  & 51.3 & 71.9 & 68.2 & 1849 & 71.7 & 84.5\% \\
				PDrop\dag  & 52.0 & 74.8 & 69.7 & 1886 & 73.6 & 86.6\% \\ 
				VisionZip\dag  & 52.4 & 78.9 & \textbf{74.1} & 2003 & \textbf{75.6} & 90.3\%    \\
				
				\rowcolor{blue!12}
				\textbf{ApET (Ours)} & \textbf{56.1} & \textbf{82.4} & \textbf{74.1} & \textbf{2055} & 72.6 & \textbf{92.1\%} \\
				
				\bottomrule[1.2pt]
			\end{tabular}
		}
	\end{table}
    
\begin{table}[t]
\centering
\caption{\footnotesize{\textbf{Performance of ApET on Video-LLaVA-7B across three video understanding tasks.} The original Video-LLaVA’s video token number is 2048, while we only retain 256 tokens for all methods.}}
%			\caption{Performance of FlowCut on video understanding tasks. The original  Video-LLaVA’s video token number is 2048, while our FlowCut only retain the 256 tokens.}
			\label{tab:main_video}
			\setlength{\tabcolsep}{0.8pt}
			\small
			\renewcommand{\arraystretch}{1.325}
			\resizebox{1.0\linewidth}{!}{
				{\begin{tabular}{@{} l |cc|cc|cc|cc @{}}
						\toprule[1.2pt]
						\multirow{2}{*}{\textbf{Method}}       & \multicolumn{2}{c|}{\textbf{TGIF}}& \multicolumn{2}{c|}{\textbf{MSVD}} & \multicolumn{2}{c|}{\textbf{MSRVTT}}  & \multicolumn{2}{c}{\textbf{Avg.}}  \\ 
						%					\cline{2-11}
						& Acc    &Score     & Acc    & Score        & Acc     & Score         & Acc    & Score    \\ \hline
						
						\textcolor{gray}{Vanilla}         & \textcolor{gray}{46.9} & \textcolor{gray}{3.34}
						& \textcolor{gray}{69.8} & \textcolor{gray}{3.91} & \textcolor{gray}{57.1} & \textcolor{gray}{3.49}   & \textcolor{gray}{100\%} &\textcolor{gray}{100\%} \\
						\hline
						\multirow{2}{*}{FastV}  &44.2 &3.29 & 60.3 &3.72 & 40.6  & 3.18 & \multirow{2}{*}{83.9\%} & \multirow{2}{*}{94.9\%} \\
						& 94.2\%&98.5\%&86.4\%&95.1\%&77.1\%&91.1\%&& \\
						\hline
						\multirow{2}{*}{SparseVLM} &45.9 &3.32 & 68.6 &3.90 & 32.9  & 3.02   & \multirow{2}{*}{84.6\%} & \multirow{2}{*}{95.2\%} \\
						&98.9\%&99.4\%&98.3\%&99.7\%&57.6\%&86.5\%&& \\
						\hline
						\multirow{2}{*}{PDrop} &40.3 &3.21 & 61.5 &3.74 & 41.8  & 3.19  & \multirow{2}{*}{82.4\%} & \multirow{2}{*}{94.4\%} \\
						&85.9\%&96.1\%&88.1\%&95.7\%&73.2\%&91.4\%&& \\
						\hline
						\multirow{2}{*}{VisionZip} &44.3 &3.29 & 65.2 &3.83 & 54.5  & 3.43  & \multirow{2}{*}{94.4\%} & \multirow{2}{*}{98.2\%} \\
						&94.5\%&98.5\%&93.4\%&98.0\%&95.4\%&98.3\%&& \\
						\hline
						
						\multirow{2}{*}{\textbf{ApET (ours)}} &\textbf{47.6} &\textbf{3.37}&\textbf{70.4} &\textbf{3.94} & \textbf{56.9}  &\textbf{3.47}  
						&\multirow{2}{*}{\textbf{100.7\%}}
						&\multirow{2}{*}{\textbf{100.4\%}} \\
						
						&101.5\% &100.9\%&100.9\%&100.8\%&99.6\%&99.4\%&& \\
						
						\bottomrule[1.2pt]
			\end{tabular}}}
\end{table}

\noindent \textbf{Video Understanding Tasks. } 
To evaluate the proposed method’s effectiveness for video understanding, we integrated it into Video-LLaVA, where each video is encoded as 8 frames of 256 visual tokens (2048 tokens in total). Under the aggressive compression budget of retaining only 256 tokens (32 per frame), ApET achieves 100.7\% averaged accuracy across the three standard benchmarks, surpassing the previous state-of-the-art VisionZip by a substantial 6.3\%, as shown in Table~\ref{tab:main_video}.
This exceptional result not only establishes a new state of the art, but also surpasses the performance of the original Video-LLaVA-7B model while preserving merely 12.5\% of the original visual tokens. Such a finding provides strong empirical support for the hypothesis that video sequences contain a considerable proportion of redundant, distracting, or even misleading tokens. Consequently, token compression methods serve a dual purpose: they markedly enhance computational efficiency and simultaneously act as a denoising mechanism that removes spurious visual cues, thereby improving the model's overall performance.

Furthermore, ApET demonstrates consistent superiority across all three video understanding benchmarks, whereas other methods display pronounced dataset-specific sensitivity. For example, VisionZip delivers competitive accuracy on TGIF and MSRVT yet incurs a performance drop on MSVD. In contrast, ApET maintains robust performance across every evaluation set, underscoring its stability and reliability in diverse video understanding scenarios.
Closer analysis reveals that the performance gain afforded by ApET is even more pronounced for video than for image understanding. We attribute this phenomenon to the exacerbated positional bias inherent in attention-based pruning strategies when they are confronted with long, temporally extended contexts. By eschewing any dependence on attention weights, ApET eliminates this bias entirely, enabling superior compression quality and, consequently, stronger generalization at the video level.

\begin{table}[t]
\centering
\caption{
    \textbf{Ablation study on sampling strategies across three image understanding tasks.} Experiments are performed with LLaVA-1.5-7B and Qwen2.5-VL-7B at token compression ratios of 11.1\% and 10\%, respectively. “Random” denotes random sampling, “DPC” indicates density peak clustering, and “FPS” refers to farthest point sampling.
    }
    \label{tab:ab_sampling}
    \footnotesize
    \setlength{\tabcolsep}{7pt}
    \resizebox{1.0\linewidth}{!}{
    \begin{tabular}{c|c|c|c|c}
        \toprule
        \footnotesize\textbf{Model} &  \footnotesize\textbf{Sampling}  &  \footnotesize\textbf{GQA}  &  \footnotesize\textbf{POPE}  & \footnotesize\textbf{MME}
        \\
        \toprule
        \multirow{3}{*}{LLaVA-1.5-7B} 
         &
        Random
        &
        55.7
        &
        82.5 
        &
        1634
        \\
        &
        DPC
        &
        \textbf{57.1}
        &
        \textbf{84.8} 
        &
        1691
        \\
        &
        FPS %
        &
        56.9
        &
        84.4
        &
        \textbf{1714}
        \\
            \midrule 
        \multirow{3}{*}{Qwen2.5-VL-7B} %
        &
        Random
        &
        55.6 
        &
        80.8 
        &
        1944
        \\
        &
        DPC
        &
        \textbf{56.5}
        &
        \textbf{82.5}
        &
        1998
        \\
        &
        FPS %
        &
        56.1
        &
        82.4
        &
        \textbf{2055}
        \\
    \bottomrule
    \end{tabular}
    }
\end{table}

\subsection{Ablation Studies}
\textbf{Effectiveness of Different Sampling Methods. } The Token Selection operation in our method identifies base tokens for the linear approximation process. In Table~\ref{tab:ab_sampling}, we evaluate the impact of three sampling strategies on ApET performance:
1) Random sampling, where $M$ visual tokens are randomly selected as base tokens; 2) Density Peak Clustering (DPC), which derives $M$ cluster centers from all visual tokens; 3) Farthest Point Sampling (FPS), which selects $M$ maximally diverse tokens as base tokens. Experimental results indicate that both FPS and DPC achieve competitive performance. However, DPC requires computing $M$ cluster centers from $N$ visual tokens, incurring significant computational overhead when $N$ is large—an inefficiency that hampers practical deployment. Thus, we adopt FPS as the default sampling strategy for ApET, striking an optimal balance between accuracy and efficiency. Notably, even under random sampling, ApET maintains robust performance, validating the efficacy of our core insight: leveraging approximation error as a proxy for token importance in attention mechanisms.

\noindent \textbf{Analysis on Hyper-parameter.} In our token selection operation, we select $M$ visual tokens as base tokens for subsequent linear approximation. To assess the impact of varying $M$ on model performance, we conducted an ablation study in Table~\ref{tab:ab_M}. Experimental results indicate that performance is stable across a wide range of $M$, peaking at $M=10$. We therefore adopt $M=10$ for all experiments.
Furthermore, we summarize two conclusions.
1) $M$ is insensitive to the performance, which indicates that our method is robust to hyper-parameter settings.
2) $M$ is vastly smaller than the total number of visual tokens $N$ ($N=576$ for LLaVA-1.5-7B), so the additional computational overhead introduced by the sampling or the approximation step is negligible.

\begin{table}[t]
\centering
\caption{
    \textbf{Ablation studies of Hyper-parameter.} Experiments are performed with LLaVA-1.5-7B and Qwen2.5-VL-7B at token compression ratios of 11.1\% and 10\%, respectively. 
    }
    \label{tab:ab_M}
    \footnotesize
    \setlength{\tabcolsep}{7pt}
    \resizebox{1.0\linewidth}{!}{
    \begin{tabular}{c|c|c|c|c}
        \toprule
        \footnotesize\textbf{Model} &  \footnotesize\textbf{Settings}  &  \footnotesize\textbf{GQA}  &  \footnotesize\textbf{POPE}  & \footnotesize\textbf{MME}
        \\
        \toprule
        \multirow{5}{*}{LLaVA-1.5-7B} 
         &
        $M=6$
        &
        56.0
        &
        83.5 
        &
        1676
        \\
        &
        $M=8$
        &
        56.7
        &
        84.1 
        &
        1705
        \\
        &
        $M=10$ %
        &
        \textbf{56.9}
        &
        \textbf{84.4}
        &
        \textbf{1714}
        \\
        &
        $M=12$ %
        &
        56.8
        &
        83.9
        &
        1699
        \\
        &
        $M=14$ %
        &
        56.4
        &
        83.7
        &
        1663
        \\
            \midrule 
        \multirow{5}{*}{Qwen2.5-VL-7B} %
        &
        $M=6$
        &
        55.9 
        &
        81.7
        &
        2014
        \\
        &
        $M=8$
        &
        56.2
        &
        82.0
        &
        2047
        \\
        &
        $M=10$ %
        &
        56.1
        &
        \textbf{82.4}
        &
        \textbf{2055}
        \\
        &
        $M=12$ %
        &
        \textbf{56.4}
        &
        82.3
        &
        2044
        \\
        &
        $M=14$ %
        &
        55.7
        &
        81.5
        &
        1972
        \\
    \bottomrule
    \end{tabular}
    }
\end{table}

\begin{table*}[t]
		\centering
		\caption{\textbf{Efficiency analysis of ApET on LLaVA-1.5-7B and Qwen2.5-VL-7B.} We report total inference time, prefilling time, and TFLOPs; \(\Delta\) indicates the speedup ratio. All measurements are performed on a single A100 GPU using the POPE benchmark.}
		\label{tab:efficiency}
		\vspace{0.2cm}
		\begin{minipage}[t]{0.49\linewidth}
			\centering
			%		\vspace{0.1cm}
			\setlength{\tabcolsep}{2.5pt}
			\renewcommand{\arraystretch}{1.2}
			\footnotesize
			\resizebox{\linewidth}{!}{
				\begin{tabular}{lcccccc}
					\toprule
					\multirow{2}{*}{\textbf{Methods}} & \multirow{2}{*}{\textbf{Token}} & 
					\textbf{Total} & 
					\multirow{2}{*}{\textbf{\(\Delta\)$\uparrow$}} & 
					\textbf{Prefilling}  & 
					\multirow{2}{*}{\textbf{\(\Delta\)$\uparrow$}} &
					\multirow{2}{*}{\textbf{TFLOPs}}\\
					& &
					\textbf{\hspace{5pt}Time$\downarrow$} &  & 
					\textbf{ Time$\downarrow$} &  &\\
					\midrule
					\textcolor{gray}{LLaVA-1.5-7B} & \textcolor{gray}{100\%} & \textcolor{gray}{14:47} & \textcolor{gray}{1.0$\times$}& \textcolor{gray}{86.9ms} & \textcolor{gray}{1.0$\times$}&\textcolor{gray}{8.82}\\
					+ SparseVLM &11.1\% & 13:32 & 1.09$\times$ & 77.8ms & 1.12$\times$ &2.31 \\
					+ PDrop &11.1\% & 11:15 & 1.31$\times$ & 69.2ms & 1.26$\times$ &2.16 \\
					+ VisionZip &11.1\% & 10:36 & 1.39$\times$ & 64.4ms & 1.35$\times$ & \textbf{2.03} \\
					\rowcolor{blue!12} %\rowcolor{cyan!10}
					\textbf{+ ApET (ours) } &11.1\% & \textbf{10:08} & \textbf{1.46$\times$} & \textbf{62.8ms} & \textbf{1.38$\times$}  & 2.09 \\
					\bottomrule
				\end{tabular}
			}
		\end{minipage}
		\hfill 
		\begin{minipage}[t]{0.49\linewidth}
			\centering
			%		\caption{Efficiency analysis of FlowCut on LLaVA-NeXT 7B}
			%		\label{tab:popenext}
			%		\vspace{0.1cm}
			\setlength{\tabcolsep}{1.85pt}
			\renewcommand{\arraystretch}{1.2}
			\footnotesize
			\resizebox{\linewidth}{!}{
				\begin{tabular}{lcccccc}
					\toprule
					\multirow{2}{*}{\textbf{Methods}}  & \multirow{2}{*}{\textbf{Token}} &
					\textbf{Total} & 
					\multirow{2}{*}{\textbf{\(\Delta\)$\uparrow$}} & 
					\textbf{Prefilling}  & 
					\multirow{2}{*}{\textbf{\(\Delta\)$\uparrow$}} &
					\multirow{2}{*}{\textbf{TFLOPs}}\\
					  & &
					\textbf{\hspace{5pt}Time$\downarrow$} &  & 
					\textbf{ Time$\downarrow$} &  &\\
					\midrule
					\textcolor{gray}{Qwen2.5-VL-7B}  & \textcolor{gray}{100\%} & \textcolor{gray}{34:14} &\hspace{2pt}\textcolor{gray}{1.0$\times$}& \textcolor{gray}{158.5ms} & \textcolor{gray}{1.0$\times$}&\textcolor{gray}{14.52}\\
					+ SparseVLM &10\% & {33:36}  &\hspace{2pt}1.02$\times$ & 136.8ms & 1.16$\times$ &5.97 \\
					+ PDrop &10\% & {32:20} &\hspace{2pt}1.06$\times$ & 125.5ms & 1.26$\times$ & 5.85 \\
					+ VisionZip &10\%  & {30:06} &\hspace{2pt}1.14$\times$ & 119.3ms & 1.33$\times$ &\textbf{5.68} \\
					\rowcolor{blue!12} %\rowcolor{cyan!10}
					\textbf{+ ApET (ours)} & 10\% & \textbf{26:18} &\hspace{2pt}\textbf{1.30$\times$} & \textbf{104.7ms} & \textbf{1.51$\times$}  &5.70\\
					\bottomrule[1.1pt]
				\end{tabular}
			}
		\end{minipage}
	\end{table*}

\begin{figure*}[t]
    \vspace{0.2cm}
    \centering
    \includegraphics[width=1.0\linewidth]{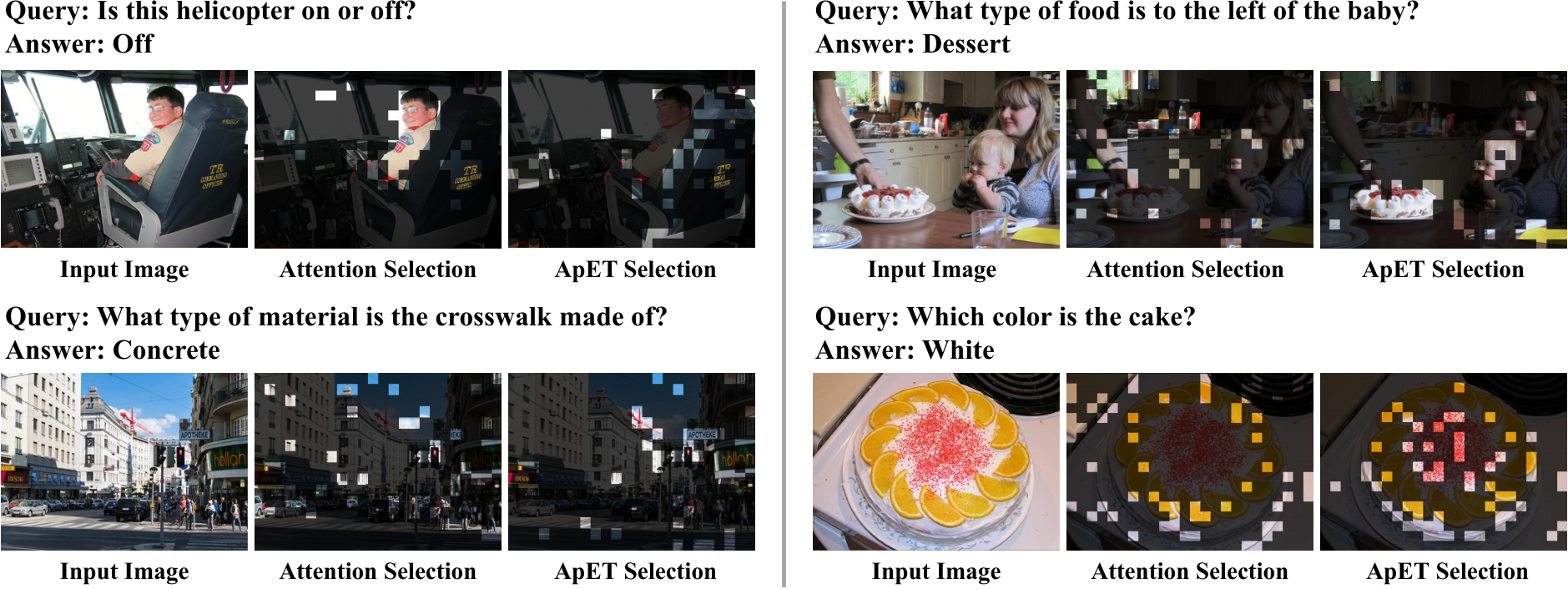}
\caption{\textbf{Visualization of token compression.} We present four representative failure cases in which attention-driven token selection misguides the final prediction. For each case, we visualize the input question and its ground-truth answer, the original image, the subset of visual tokens preserved when ranking by attention weights, and the subset preserved by the approximation-error criterion proposed in this work.
The retained tokens are highlighted for clear comparison.}
    \label{fig:vis}
    \vspace{-0.2cm}
\end{figure*}
    
\noindent \textbf{Efficiency Analysis.} Our proposed ApET significantly enhances inference efficiency in VLMs while maintaining performance on par with existing approaches. As shown in Table~\ref{tab:efficiency}, we evaluate total inference time, prefilling time, and FLOPs on both LLaVA-1.5-7B and Qwen2.5-VL-7B, comparing ApET against several state-of-the-art methods.
ApET consistently achieves optimal efficiency across both model architectures. 
Specifically, it yields a 1.46$\times$ speedup in total inference time and a 1.38× speedup in the prefilling time on LLaVA-1.5-7B.
Notably, while competing methods yield efficiency improvements on LLaVA-1.5-7B, they underperform the baseline model on the more advanced Qwen2.5-VL-7B. This degradation stems from their reliance on attention weights to assess token importance. While LLaVA supports the extraction of these weights, Qwen2.5-VL does not, necessitating a costly recomputation of attention scores. This overhead introduces significant computational burden, ultimately reducing overall efficiency.
In contrast, ApET estimates token importance via linear approximations applied directly to token representations, obviating any dependence on model internal intermediate results. This design enables seamless adaptation to architectures such as Qwen2.5-VL without additional compute. 
These results validate our central contribution: a model-agnostic token compression mechanism. We argue that effective compression strategies should be decoupled from model-specific internals, eschewing reliance on any architecture dependent computations—to ensure robust generalization across diverse VLMs.

\noindent \textbf{Visualization.} 
To further illustrate the effectiveness of our proposed ApET method, we provide qualitative visualizations of token selection results in Figure~\ref{fig:vis}. These examples are drawn from the GQA dataset and showcase representative cases where attention-based token pruning methods fail to preserve task-relevant visual information, leading to incorrect predictions. In contrast, ApET guided by approximation error, successfully retains informative tokens and produces accurate answers.
Each visualization includes the input image, the textual question, the ground-truth answer, and two sets of preserved tokens: one selected by attention weight based ranking (left), and the other by our approximation-error criterion (right). The retained tokens are highlighted on the image for clarity.
For instance, in the first example, the query asks whether a helicopter is "on or off". Attention-based pruning fails to retain visual tokens corresponding to the cockpit indicators, resulting in an incorrect prediction. ApET, however, preserves tokens that capture the static state of the helicopter, enabling the model to infer the correct answer.

These visualizations underscore a key insight: approximation error serves as a more content-aware metric for token importance, unlike attention weights which are susceptible to positional and contextual biases. By focusing on reconstructability, ApET inherently prioritizes tokens that carry distinctive and irreplaceable visual information, leading to more robust and interpretable token reduction.

\section{Conclusion}
Vision-language models (VLMs) demand prohibitive computational overhead because of the massive number of visual tokens. While existing token compression methods that rely on attention weights not only introduce positional bias but also fail to be compatible with efficient attention kernels.
In this work, we present ApET, an attention-free token compression framework that leverages approximation error to evaluate visual token importance from an information-theoretic perspective. By reconstructing tokens via linear approximation and retaining those with high reconstruction error, ApET preserves informative content while eliminating positional bias inherent in attention-based methods. Extensive experiments demonstrate that ApET achieves strong performance across both image and video understanding benchmarks, even under aggressive compression rates. Notably, it is fully compatible with efficient attention implementations like FlashAttention, enabling practical acceleration in real-world deployments.

\newpage
{
    \small
    \bibliographystyle{ieeenat_fullname}
    \bibliography{main}

\begin{thebibliography}{59}
\providecommand{\natexlab}[1]{#1}
\providecommand{\url}[1]{\texttt{#1}}
\expandafter\ifx\csname urlstyle\endcsname\relax
  \providecommand{\doi}[1]{doi: #1}\else
  \providecommand{\doi}{doi: \begingroup \urlstyle{rm}\Url}\fi

\bibitem[Achiam et~al.(2023)Achiam, Adler, Agarwal, Ahmad, Akkaya, Aleman, Almeida, Altenschmidt, Altman, Anadkat, et~al.]{achiam2023gpt}
Josh Achiam, Steven Adler, Sandhini Agarwal, Lama Ahmad, Ilge Akkaya, Florencia~Leoni Aleman, Diogo Almeida, Janko Altenschmidt, Sam Altman, Shyamal Anadkat, et~al.
\newblock Gpt-4 technical report.
\newblock \emph{arXiv preprint arXiv:2303.08774}, 2023.

\bibitem[Alayrac et~al.(2022)Alayrac, Donahue, Luc, Miech, Barr, Hasson, Lenc, Mensch, Millican, Reynolds, et~al.]{alayrac2022flamingo}
Jean-Baptiste Alayrac, Jeff Donahue, Pauline Luc, Antoine Miech, Iain Barr, Yana Hasson, Karel Lenc, Arthur Mensch, Katherine Millican, Malcolm Reynolds, et~al.
\newblock Flamingo: a visual language model for few-shot learning.
\newblock \emph{Advances in neural information processing systems}, 35:\penalty0 23716--23736, 2022.

\bibitem[Arif et~al.(2025)Arif, Yoon, Nikolopoulos, Vandierendonck, John, and Ji]{arif2025hired}
Kazi Hasan~Ibn Arif, JinYi Yoon, Dimitrios~S Nikolopoulos, Hans Vandierendonck, Deepu John, and Bo Ji.
\newblock Hired: Attention-guided token dropping for efficient inference of high-resolution vision-language models.
\newblock In \emph{Proceedings of the AAAI Conference on Artificial Intelligence}, pages 1773--1781, 2025.

\bibitem[Bai et~al.(2023)Bai, Bai, Yang, Wang, Tan, Wang, Lin, Zhou, and Zhou]{bai2023qwen}
Jinze Bai, Shuai Bai, Shusheng Yang, Shijie Wang, Sinan Tan, Peng Wang, Junyang Lin, Chang Zhou, and Jingren Zhou.
\newblock Qwen-vl: A frontier large vision-language model with versatile abilities.
\newblock \emph{arXiv preprint arXiv:2308.12966}, 1\penalty0 (2):\penalty0 3, 2023.

\bibitem[Bai et~al.(2025)Bai, Chen, Liu, Wang, Ge, Song, Dang, Wang, Wang, Tang, et~al.]{bai2025qwen2}
Shuai Bai, Keqin Chen, Xuejing Liu, Jialin Wang, Wenbin Ge, Sibo Song, Kai Dang, Peng Wang, Shijie Wang, Jun Tang, et~al.
\newblock Qwen2. 5-vl technical report.
\newblock \emph{arXiv preprint arXiv:2502.13923}, 2025.

\bibitem[Bolya et~al.(2022)Bolya, Fu, Dai, Zhang, Feichtenhofer, and Hoffman]{bolya2022token}
Daniel Bolya, Cheng-Yang Fu, Xiaoliang Dai, Peizhao Zhang, Christoph Feichtenhofer, and Judy Hoffman.
\newblock Token merging: Your vit but faster.
\newblock \emph{arXiv preprint arXiv:2210.09461}, 2022.

\bibitem[Brauwers and Frasincar(2021)]{brauwers2021general}
Gianni Brauwers and Flavius Frasincar.
\newblock A general survey on attention mechanisms in deep learning.
\newblock \emph{IEEE transactions on knowledge and data engineering}, 35\penalty0 (4):\penalty0 3279--3298, 2021.

\bibitem[Chen et~al.(2023)Chen, Zhang, Zeng, Zhang, Zhu, and Zhao]{chen2023shikra}
Keqin Chen, Zhao Zhang, Weili Zeng, Richong Zhang, Feng Zhu, and Rui Zhao.
\newblock Shikra: Unleashing multimodal llm's referential dialogue magic.
\newblock \emph{arXiv preprint arXiv:2306.15195}, 2023.

\bibitem[Chen et~al.(2024{\natexlab{a}})Chen, Li, Dong, Zhang, He, Wang, Zhao, and Lin]{chen2024sharegpt4v}
Lin Chen, Jinsong Li, Xiaoyi Dong, Pan Zhang, Conghui He, Jiaqi Wang, Feng Zhao, and Dahua Lin.
\newblock Sharegpt4v: Improving large multi-modal models with better captions.
\newblock In \emph{European Conference on Computer Vision}, pages 370--387. Springer, 2024{\natexlab{a}}.

\bibitem[Chen et~al.(2024{\natexlab{b}})Chen, Zhao, Liu, Bai, Lin, Zhou, and Chang]{chen2024image}
Liang Chen, Haozhe Zhao, Tianyu Liu, Shuai Bai, Junyang Lin, Chang Zhou, and Baobao Chang.
\newblock An image is worth 1/2 tokens after layer 2: Plug-and-play inference acceleration for large vision-language models.
\newblock In \emph{European Conference on Computer Vision}, pages 19--35. Springer, 2024{\natexlab{b}}.

\bibitem[Chen et~al.(2024{\natexlab{c}})Chen, Xue, Li, Hu, Zhu, Li, Fang, Tang, Yang, Liu, et~al.]{chen2024longvila}
Yukang Chen, Fuzhao Xue, Dacheng Li, Qinghao Hu, Ligeng Zhu, Xiuyu Li, Yunhao Fang, Haotian Tang, Shang Yang, Zhijian Liu, et~al.
\newblock Longvila: Scaling long-context visual language models for long videos.
\newblock \emph{arXiv preprint arXiv:2408.10188}, 2024{\natexlab{c}}.

\bibitem[Chen et~al.(2024{\natexlab{d}})Chen, Wu, Wang, Su, Chen, Xing, Zhong, Zhang, Zhu, Lu, et~al.]{chen2024internvl}
Zhe Chen, Jiannan Wu, Wenhai Wang, Weijie Su, Guo Chen, Sen Xing, Muyan Zhong, Qinglong Zhang, Xizhou Zhu, Lewei Lu, et~al.
\newblock Internvl: Scaling up vision foundation models and aligning for generic visual-linguistic tasks.
\newblock In \emph{Proceedings of the IEEE/CVF conference on computer vision and pattern recognition}, pages 24185--24198, 2024{\natexlab{d}}.

\bibitem[Child et~al.(2019)Child, Gray, Radford, and Sutskever]{child2019generating}
Rewon Child, Scott Gray, Alec Radford, and Ilya Sutskever.
\newblock Generating long sequences with sparse transformers.
\newblock \emph{arXiv preprint arXiv:1904.10509}, 2019.

\bibitem[Chung et~al.(2024)Chung, Hou, Longpre, Zoph, Tay, Fedus, Li, Wang, Dehghani, Brahma, et~al.]{chung2024scaling}
Hyung~Won Chung, Le Hou, Shayne Longpre, Barret Zoph, Yi Tay, William Fedus, Yunxuan Li, Xuezhi Wang, Mostafa Dehghani, Siddhartha Brahma, et~al.
\newblock Scaling instruction-finetuned language models.
\newblock \emph{Journal of Machine Learning Research}, 25\penalty0 (70):\penalty0 1--53, 2024.

\bibitem[Dai et~al.(2023)Dai, Li, Li, Tiong, Zhao, Wang, Li, Fung, and Hoi]{dai2023instructblip}
Wenliang Dai, Junnan Li, Dongxu Li, Anthony Tiong, Junqi Zhao, Weisheng Wang, Boyang Li, Pascale~N Fung, and Steven Hoi.
\newblock Instructblip: Towards general-purpose vision-language models with instruction tuning.
\newblock \emph{Advances in neural information processing systems}, 36:\penalty0 49250--49267, 2023.

\bibitem[Dao(2023)]{dao2023flashattention}
Tri Dao.
\newblock Flashattention-2: Faster attention with better parallelism and work partitioning.
\newblock \emph{arXiv preprint arXiv:2307.08691}, 2023.

\bibitem[Dao et~al.(2022)Dao, Fu, Ermon, Rudra, and R{\'e}]{dao2022flashattention}
Tri Dao, Dan Fu, Stefano Ermon, Atri Rudra, and Christopher R{\'e}.
\newblock Flashattention: Fast and memory-efficient exact attention with io-awareness.
\newblock \emph{Advances in neural information processing systems}, 35:\penalty0 16344--16359, 2022.

\bibitem[Jaegle et~al.(2021)Jaegle, Gimeno, Brock, Vinyals, Zisserman, and Carreira]{jaegle2021perceiver}
Andrew Jaegle, Felix Gimeno, Andy Brock, Oriol Vinyals, Andrew Zisserman, and Joao Carreira.
\newblock Perceiver: General perception with iterative attention.
\newblock In \emph{International conference on machine learning}, pages 4651--4664. PMLR, 2021.

\bibitem[Jia et~al.(2024)Jia, Yu, Ma, Fang, Zhang, Ouyang, Zhang, Yu, and Jiang]{jia2024leopard}
Mengzhao Jia, Wenhao Yu, Kaixin Ma, Tianqing Fang, Zhihan Zhang, Siru Ouyang, Hongming Zhang, Dong Yu, and Meng Jiang.
\newblock Leopard: A vision language model for text-rich multi-image tasks.
\newblock \emph{arXiv preprint arXiv:2410.01744}, 2024.

\bibitem[Kim et~al.(2024)Kim, Pertsch, Karamcheti, Xiao, Balakrishna, Nair, Rafailov, Foster, Lam, Sanketi, et~al.]{kim2024openvla}
Moo~Jin Kim, Karl Pertsch, Siddharth Karamcheti, Ted Xiao, Ashwin Balakrishna, Suraj Nair, Rafael Rafailov, Ethan Foster, Grace Lam, Pannag Sanketi, et~al.
\newblock Openvla: An open-source vision-language-action model.
\newblock \emph{arXiv preprint arXiv:2406.09246}, 2024.

\bibitem[Kondratyuk et~al.(2021)Kondratyuk, Yuan, Li, Zhang, Tan, Brown, and Gong]{kondratyuk2021movinets}
Dan Kondratyuk, Liangzhe Yuan, Yandong Li, Li Zhang, Mingxing Tan, Matthew Brown, and Boqing Gong.
\newblock Movinets: Mobile video networks for efficient video recognition.
\newblock In \emph{Proceedings of the IEEE/CVF conference on computer vision and pattern recognition}, pages 16020--16030, 2021.

\bibitem[Li et~al.(2024{\natexlab{a}})Li, Zhang, Guo, Zhang, Li, Zhang, Zhang, Zhang, Li, Liu, et~al.]{li2024llava}
Bo Li, Yuanhan Zhang, Dong Guo, Renrui Zhang, Feng Li, Hao Zhang, Kaichen Zhang, Peiyuan Zhang, Yanwei Li, Ziwei Liu, et~al.
\newblock Llava-onevision: Easy visual task transfer.
\newblock \emph{arXiv preprint arXiv:2408.03326}, 2024{\natexlab{a}}.

\bibitem[Li et~al.(2023)Li, Li, Savarese, and Hoi]{li2023blip}
Junnan Li, Dongxu Li, Silvio Savarese, and Steven Hoi.
\newblock Blip-2: Bootstrapping language-image pre-training with frozen image encoders and large language models.
\newblock In \emph{International conference on machine learning}, pages 19730--19742. PMLR, 2023.

\bibitem[Li et~al.(2024{\natexlab{b}})Li, Zhang, Wang, Zhong, Chen, Chu, Liu, and Jia]{li2024mini}
Yanwei Li, Yuechen Zhang, Chengyao Wang, Zhisheng Zhong, Yixin Chen, Ruihang Chu, Shaoteng Liu, and Jiaya Jia.
\newblock Mini-gemini: Mining the potential of multi-modality vision language models.
\newblock \emph{arXiv preprint arXiv:2403.18814}, 2024{\natexlab{b}}.

\bibitem[Lin et~al.(2023)Lin, Ye, Zhu, Cui, Ning, Jin, and Yuan]{lin2023video}
Bin Lin, Yang Ye, Bin Zhu, Jiaxi Cui, Munan Ning, Peng Jin, and Li Yuan.
\newblock Video-llava: Learning united visual representation by alignment before projection.
\newblock \emph{arXiv preprint arXiv:2311.10122}, 2023.

\bibitem[Liu et~al.(2024{\natexlab{a}})Liu, Feng, Xue, Wang, Wu, Lu, Zhao, Deng, Zhang, Ruan, et~al.]{liu2024deepseek}
Aixin Liu, Bei Feng, Bing Xue, Bingxuan Wang, Bochao Wu, Chengda Lu, Chenggang Zhao, Chengqi Deng, Chenyu Zhang, Chong Ruan, et~al.
\newblock Deepseek-v3 technical report.
\newblock \emph{arXiv preprint arXiv:2412.19437}, 2024{\natexlab{a}}.

\bibitem[Liu et~al.(2024{\natexlab{b}})Liu, Li, Li, and Lee]{liu2024improved}
Haotian Liu, Chunyuan Li, Yuheng Li, and Yong~Jae Lee.
\newblock Improved baselines with visual instruction tuning.
\newblock In \emph{Proceedings of the IEEE/CVF conference on computer vision and pattern recognition}, pages 26296--26306, 2024{\natexlab{b}}.

\bibitem[Liu et~al.(2024{\natexlab{c}})Liu, Li, Li, Li, Zhang, Shen, and Lee]{liu2024llavanext}
Haotian Liu, Chunyuan Li, Yuheng Li, Bo Li, Yuanhan Zhang, Sheng Shen, and Yong~Jae Lee.
\newblock Llavanext: Improved reasoning, ocr, and world knowledge, 2024{\natexlab{c}}.

\bibitem[Liu et~al.(2024{\natexlab{d}})Liu, Shi, Hong, Hu, Yin, and Zhang]{liu2024multi}
Ting Liu, Liangtao Shi, Richang Hong, Yue Hu, Quanjun Yin, and Linfeng Zhang.
\newblock Multi-stage vision token dropping: Towards efficient multimodal large language model.
\newblock \emph{arXiv preprint arXiv:2411.10803}, 2024{\natexlab{d}}.

\bibitem[Lu et~al.(2025{\natexlab{a}})Lu, Yu, Lu, Rajan, Ng, Kot, and Jiang]{lu2025mambatad}
Hui Lu, Yi Yu, Shijian Lu, Deepu Rajan, Boon~Poh Ng, Alex~C Kot, and Xudong Jiang.
\newblock Mambatad: When state-space models meet long-range temporal action detection.
\newblock \emph{arXiv preprint arXiv:2511.17929}, 2025{\natexlab{a}}.

\bibitem[Lu et~al.(2025{\natexlab{b}})Lu, Yu, Xia, Yang, Rajan, Ng, Kot, and Jiang]{lu2025pretrain}
Hui Lu, Yi Yu, Song Xia, Yiming Yang, Deepu Rajan, Boon~Poh Ng, Alex Kot, and Xudong Jiang.
\newblock From pretrain to pain: Adversarial vulnerability of video foundation models without task knowledge.
\newblock \emph{arXiv preprint arXiv:2511.07049}, 2025{\natexlab{b}}.

\bibitem[Ma et~al.(2024)Ma, Jin, Wang, Xian, Feng, and Yang]{ma2024vista}
Fan Ma, Xiaojie Jin, Heng Wang, Yuchen Xian, Jiashi Feng, and Yi Yang.
\newblock Vista-llama: Reducing hallucination in video language models via equal distance to visual tokens.
\newblock In \emph{Proceedings of the IEEE/CVF Conference on Computer Vision and Pattern Recognition}, pages 13151--13160, 2024.

\bibitem[Maaz et~al.(2023)Maaz, Rasheed, Khan, and Khan]{maaz2023video}
Muhammad Maaz, Hanoona Rasheed, Salman Khan, and Fahad~Shahbaz Khan.
\newblock Video-chatgpt: Towards detailed video understanding via large vision and language models.
\newblock \emph{arXiv preprint arXiv:2306.05424}, 2023.

\bibitem[Mehta and Rastegari(2021)]{mehta2021mobilevit}
Sachin Mehta and Mohammad Rastegari.
\newblock Mobilevit: light-weight, general-purpose, and mobile-friendly vision transformer.
\newblock \emph{arXiv preprint arXiv:2110.02178}, 2021.

\bibitem[Qu et~al.(2025)Qu, Chen, Wei, Lin, Chen, and Huang]{qu2025mobile}
Guanqiao Qu, Qiyuan Chen, Wei Wei, Zheng Lin, Xianhao Chen, and Kaibin Huang.
\newblock Mobile edge intelligence for large language models: A contemporary survey.
\newblock \emph{IEEE Communications Surveys \& Tutorials}, 2025.

\bibitem[Shang et~al.(2024)Shang, Cai, Xu, Lee, and Yan]{shang2024llava}
Yuzhang Shang, Mu Cai, Bingxin Xu, Yong~Jae Lee, and Yan Yan.
\newblock Llava-prumerge: Adaptive token reduction for efficient large multimodal models.
\newblock \emph{arXiv preprint arXiv:2403.15388}, 2024.

\bibitem[Shannon(1948)]{shannon1948mathematical}
Claude~E Shannon.
\newblock A mathematical theory of communication.
\newblock \emph{The Bell system technical journal}, 27\penalty0 (3):\penalty0 379--423, 1948.

\bibitem[Tan et~al.(2025)Tan, Ye, Tu, Cao, Yang, Zhang, Zhou, and Chen]{tan2025tokencarve}
Xudong Tan, Peng Ye, Chongjun Tu, Jianjian Cao, Yaoxin Yang, Lin Zhang, Dongzhan Zhou, and Tao Chen.
\newblock Tokencarve: Information-preserving visual token compression in multimodal large language models.
\newblock \emph{arXiv preprint arXiv:2503.10501}, 2025.

\bibitem[Team et~al.(2024)Team, Georgiev, Lei, Burnell, Bai, Gulati, Tanzer, Vincent, Pan, Wang, et~al.]{team2024gemini}
Gemini Team, Petko Georgiev, Ving~Ian Lei, Ryan Burnell, Libin Bai, Anmol Gulati, Garrett Tanzer, Damien Vincent, Zhufeng Pan, Shibo Wang, et~al.
\newblock Gemini 1.5: Unlocking multimodal understanding across millions of tokens of context.
\newblock \emph{arXiv preprint arXiv:2403.05530}, 2024.

\bibitem[Tong et~al.(2024)Tong, Brown, Wu, Woo, IYER, Akula, Yang, Yang, Middepogu, Wang, et~al.]{tong2024cambrian}
Peter Tong, Ellis Brown, Penghao Wu, Sanghyun Woo, Adithya Jairam~Vedagiri IYER, Sai~Charitha Akula, Shusheng Yang, Jihan Yang, Manoj Middepogu, Ziteng Wang, et~al.
\newblock Cambrian-1: A fully open, vision-centric exploration of multimodal llms.
\newblock \emph{Advances in Neural Information Processing Systems}, 37:\penalty0 87310--87356, 2024.

\bibitem[Touvron et~al.(2023)Touvron, Lavril, Izacard, Martinet, Lachaux, Lacroix, Rozi{\`e}re, Goyal, Hambro, Azhar, et~al.]{touvron2023llama}
Hugo Touvron, Thibaut Lavril, Gautier Izacard, Xavier Martinet, Marie-Anne Lachaux, Timoth{\'e}e Lacroix, Baptiste Rozi{\`e}re, Naman Goyal, Eric Hambro, Faisal Azhar, et~al.
\newblock Llama: Open and efficient foundation language models.
\newblock \emph{arXiv preprint arXiv:2302.13971}, 2023.

\bibitem[Vaswani et~al.(2017)Vaswani, Shazeer, Parmar, Uszkoreit, Jones, Gomez, Kaiser, and Polosukhin]{vaswani2017attention}
Ashish Vaswani, Noam Shazeer, Niki Parmar, Jakob Uszkoreit, Llion Jones, Aidan~N Gomez, {\L}ukasz Kaiser, and Illia Polosukhin.
\newblock Attention is all you need.
\newblock \emph{Advances in neural information processing systems}, 30, 2017.

\bibitem[Wang et~al.(2025)Wang, Yu, Spadaro, Ju, Qu{\'e}tu, Xiao, and Tartaglione]{wang2025folder}
Haicheng Wang, Zhemeng Yu, Gabriele Spadaro, Chen Ju, Victor Qu{\'e}tu, Shuai Xiao, and Enzo Tartaglione.
\newblock Folder: Accelerating multi-modal large language models with enhanced performance.
\newblock \emph{arXiv preprint arXiv:2501.02430}, 2025.

\bibitem[Wen et~al.(2025)Wen, Gao, Wang, Zhang, Zhang, Li, He, and Zhang]{wen2025stop}
Zichen Wen, Yifeng Gao, Shaobo Wang, Junyuan Zhang, Qintong Zhang, Weijia Li, Conghui He, and Linfeng Zhang.
\newblock Stop looking for important tokens in multimodal language models: Duplication matters more.
\newblock \emph{arXiv preprint arXiv:2502.11494}, 2025.

\bibitem[Xia et~al.(2025)Xia, Yu, Yang, Ding, Chen, Duan, Kot, and Jiang]{xia2025theoretical}
Song Xia, Yi Yu, Wenhan Yang, Meiwen Ding, Zhuo Chen, Ling-Yu Duan, Alex~C Kot, and Xudong Jiang.
\newblock Theoretical insights in model inversion robustness and conditional entropy maximization for collaborative inference systems.
\newblock In \emph{Proceedings of the Computer Vision and Pattern Recognition Conference}, pages 8753--8763, 2025.

\bibitem[Xing et~al.(2024)Xing, Huang, Dong, Lu, Zhang, Zang, Cao, He, Wang, Wu, et~al.]{xing2024pyramiddrop}
Long Xing, Qidong Huang, Xiaoyi Dong, Jiajie Lu, Pan Zhang, Yuhang Zang, Yuhang Cao, Conghui He, Jiaqi Wang, Feng Wu, et~al.
\newblock Pyramiddrop: Accelerating your large vision-language models via pyramid visual redundancy reduction.
\newblock \emph{arXiv preprint arXiv:2410.17247}, 2024.

\bibitem[Xing et~al.(2025)Xing, Huang, Dong, Lu, Zhang, Zang, Cao, He, Wang, Wu, et~al.]{xing2025conical}
Long Xing, Qidong Huang, Xiaoyi Dong, Jiajie Lu, Pan Zhang, Yuhang Zang, Yuhang Cao, Conghui He, Jiaqi Wang, Feng Wu, et~al.
\newblock Conical visual concentration for efficient large vision-language models.
\newblock In \emph{Proceedings of the Computer Vision and Pattern Recognition Conference}, pages 14593--14603, 2025.

\bibitem[Xu et~al.(2025)Xu, Wang, Luo, and Du]{xu2025rethinking}
Rui Xu, Yunke Wang, Yong Luo, and Bo Du.
\newblock Rethinking visual token reduction in lvlms under cross-modal misalignment.
\newblock \emph{arXiv preprint arXiv:2506.22283}, 2025.

\bibitem[Yang et~al.(2024)Yang, Tian, Jiang, and Jia]{yang2024unified}
Senqiao Yang, Zhuotao Tian, Li Jiang, and Jiaya Jia.
\newblock Unified language-driven zero-shot domain adaptation.
\newblock In \emph{Proceedings of the IEEE/CVF conference on computer vision and pattern recognition}, pages 23407--23415, 2024.

\bibitem[Yang et~al.(2025{\natexlab{a}})Yang, Chen, Tian, Wang, Li, Yu, and Jia]{yang2025visionzip}
Senqiao Yang, Yukang Chen, Zhuotao Tian, Chengyao Wang, Jingyao Li, Bei Yu, and Jiaya Jia.
\newblock Visionzip: Longer is better but not necessary in vision language models.
\newblock In \emph{Proceedings of the Computer Vision and Pattern Recognition Conference}, pages 19792--19802, 2025{\natexlab{a}}.

\bibitem[Yang et~al.(2025{\natexlab{b}})Yang, Liu, Zhang, Pan, Guo, Li, Chen, Gao, Li, Guo, et~al.]{yang2025lidar}
Senqiao Yang, Jiaming Liu, Renrui Zhang, Mingjie Pan, Ziyu Guo, Xiaoqi Li, Zehui Chen, Peng Gao, Hongsheng Li, Yandong Guo, et~al.
\newblock Lidar-llm: Exploring the potential of large language models for 3d lidar understanding.
\newblock In \emph{Proceedings of the AAAI Conference on Artificial Intelligence}, pages 9247--9255, 2025{\natexlab{b}}.

\bibitem[Yao et~al.(2024)Yao, Yu, Zhang, Wang, Cui, Zhu, Cai, Li, Zhao, He, et~al.]{yao2024minicpm}
Yuan Yao, Tianyu Yu, Ao Zhang, Chongyi Wang, Junbo Cui, Hongji Zhu, Tianchi Cai, Haoyu Li, Weilin Zhao, Zhihui He, et~al.
\newblock Minicpm-v: A gpt-4v level mllm on your phone.
\newblock \emph{arXiv preprint arXiv:2408.01800}, 2024.

\bibitem[Ye et~al.(2025)Ye, Gan, Ge, Zhang, and Tang]{ye2025atp}
Xubing Ye, Yukang Gan, Yixiao Ge, Xiao-Ping Zhang, and Yansong Tang.
\newblock Atp-llava: Adaptive token pruning for large vision language models.
\newblock In \emph{Proceedings of the Computer Vision and Pattern Recognition Conference}, pages 24972--24982, 2025.

\bibitem[Zaheer et~al.(2020)Zaheer, Guruganesh, Dubey, Ainslie, Alberti, Ontanon, Pham, Ravula, Wang, Yang, et~al.]{zaheer2020big}
Manzil Zaheer, Guru Guruganesh, Kumar~Avinava Dubey, Joshua Ainslie, Chris Alberti, Santiago Ontanon, Philip Pham, Anirudh Ravula, Qifan Wang, Li Yang, et~al.
\newblock Big bird: Transformers for longer sequences.
\newblock \emph{Advances in neural information processing systems}, 33:\penalty0 17283--17297, 2020.

\bibitem[Zhang et~al.(2025{\natexlab{a}})Zhang, Ma, Fang, Yu, Zhang, Zhang, Xie, Sycara, Mi, and Yu]{zhang2025vscan}
Ce Zhang, Kaixin Ma, Tianqing Fang, Wenhao Yu, Hongming Zhang, Zhisong Zhang, Yaqi Xie, Katia Sycara, Haitao Mi, and Dong Yu.
\newblock Vscan: Rethinking visual token reduction for efficient large vision-language models.
\newblock \emph{arXiv preprint arXiv:2505.22654}, 2025{\natexlab{a}}.

\bibitem[Zhang et~al.(2025{\natexlab{b}})Zhang, Wan, Kan, Ma, Stepputtis, Ramanan, Salakhutdinov, Morency, Sycara, and Xie]{zhang2025self}
Ce Zhang, Zifu Wan, Zhehan Kan, Martin~Q Ma, Simon Stepputtis, Deva Ramanan, Russ Salakhutdinov, Louis-Philippe Morency, Katia Sycara, and Yaqi Xie.
\newblock Self-correcting decoding with generative feedback for mitigating hallucinations in large vision-language models.
\newblock \emph{arXiv preprint arXiv:2502.06130}, 2025{\natexlab{b}}.

\bibitem[Zhang et~al.(2025{\natexlab{c}})Zhang, Yao, Pi, Liang, and Fung]{zhang2025vlm2}
Jianshu Zhang, Dongyu Yao, Renjie Pi, Paul~Pu Liang, and Yi~R Fung.
\newblock Vlm2-bench: A closer look at how well vlms implicitly link explicit matching visual cues.
\newblock \emph{arXiv preprint arXiv:2502.12084}, 2025{\natexlab{c}}.

\bibitem[Zhang et~al.(2024{\natexlab{a}})Zhang, Cheng, Lu, Zhuo, Wang, Cao, Guo, She, and Zhang]{zhang2024cls}
Qizhe Zhang, Aosong Cheng, Ming Lu, Zhiyong Zhuo, Minqi Wang, Jiajun Cao, Shaobo Guo, Qi She, and Shanghang Zhang.
\newblock [cls] attention is all you need for training-free visual token pruning: Make vlm inference faster.
\newblock \emph{arXiv e-prints}, pages arXiv--2412, 2024{\natexlab{a}}.

\bibitem[Zhang et~al.(2024{\natexlab{b}})Zhang, Fan, Ma, Zheng, Huang, Cheng, Gudovskiy, Okuno, Nakata, Keutzer, et~al.]{zhang2024sparsevlm}
Yuan Zhang, Chun-Kai Fan, Junpeng Ma, Wenzhao Zheng, Tao Huang, Kuan Cheng, Denis Gudovskiy, Tomoyuki Okuno, Yohei Nakata, Kurt Keutzer, et~al.
\newblock Sparsevlm: Visual token sparsification for efficient vision-language model inference.
\newblock \emph{arXiv preprint arXiv:2410.04417}, 2024{\natexlab{b}}.

\end{thebibliography}

}

\end{document}